\documentclass[letterpaper, 10 pt, conference]{ieeeconf}  

\IEEEoverridecommandlockouts                              

\usepackage{graphicx}
\graphicspath{{figs/}}
\usepackage{bm}
\usepackage{amsmath}
\usepackage{amssymb}
\usepackage{algorithm}
\usepackage{algpseudocode}
\usepackage{times}

\newcommand{\secref}[1]{Section \ref{#1}}
\newcommand{\subsecref}[1]{Subsection \ref{#1}}

\newcommand{\figref}[1]{{Fig. \ref{#1}}}
\newcommand{\equref}[1]{{Eq. \ref{#1}}}

\title{\LARGE \bf
  Dynamic Manipulation of Flexible Objects with Torque Sequence\\ Using a Deep Neural Network
}

\author{Kento Kawaharazuka$^1$, Toru Ogawa$^2$, Juntaro Tamura$^2$, and Cota Nabeshima$^2$
  \thanks{$^1$ An author is associated with Department of Mechano-Informatics, Graduate School of Information Science and Technology, The University of Tokyo. 
    \texttt\small kawaharazuka@jsk.t.u-tokyo.ac.jp
  }
    \thanks{$^2$ Authors are associated with Preferred Networks, Inc. %
      \texttt\small \{ogawa, tamura, cota\}@preferred.jp
    }
}

\begin{document}

\maketitle
\thispagestyle{empty}
\pagestyle{empty}

\begin{abstract}
  For dynamic manipulation of flexible objects, we propose an acquisition method of a flexible object motion equation model using a deep neural network and a control method to realize a target state by calculating an optimized time-series joint torque command.
  By using the proposed method, any physics model of a target object is not needed, and the object can be controlled as intended.
  We applied this method to manipulations of a rigid object, a flexible object with and without environmental contact, and a cloth, and verified its effectiveness.
\end{abstract}

\section{INTRODUCTION}\label{sec:introduction}
  Flexible object manipulation is one of the most challenging manipulations, and has been studied vigorously.
  This type of manipulation includes knotting a rope, folding a cloth, etc., and they are fundamental motions in our daily lives.

  When we classify these manipulations from the viewpoint of static or dynamic manipulations and controls using physics-based modeling or machine learning, the studies have mainly focused on the static manipulation of flexible objects using physics-based modeling \cite{jimenez2012survey}.
  These methods estimate a geometrical relationship of an object, and determine the next movements to follow a target geometrical relationship \cite{inaba1987inaba, saha2007manipulation, elbrechter2012folding, caldwell2014optimal}.
  Meanwhile, several methods, which acquire flexible object manipulation from trial and error using a machine learning approach, have been developed \cite{lee2015learning, hu2018three, tanaka2018emd}.
  These approaches do not need any physics models of a target object, and are versatile because the models are constructed from motion data.
  On the other hand, dynamic manipulations of flexible objects such as cloth folding, knotting, etc. have been studied mostly in physics-based modeling approaches \cite{yamakawa2010motion, yamakawa2011motion}; nevertheless, there are a few studies using machine learning approaches, especially deep learning \cite{yang2017repeatable}.
  However, their target states are achieved by quasi-static movements, and they only use joint angles or joint velocities as the control input, though joint torques are critical in realizing dynamic movement.
  Also, with the spread of torque controlled robots \cite{englsberger2014overview, hyon2017design, semini2017design}, which can contact the environment softly, we should develop a dynamic manipulation method with torque command.
  In this study, to address these problems, we propose a dynamic control method of flexible objects to realize a target state from a current state using machine learning with joint torque command.

\begin{figure}[t]
  \centering
  \includegraphics[width=0.9\columnwidth]{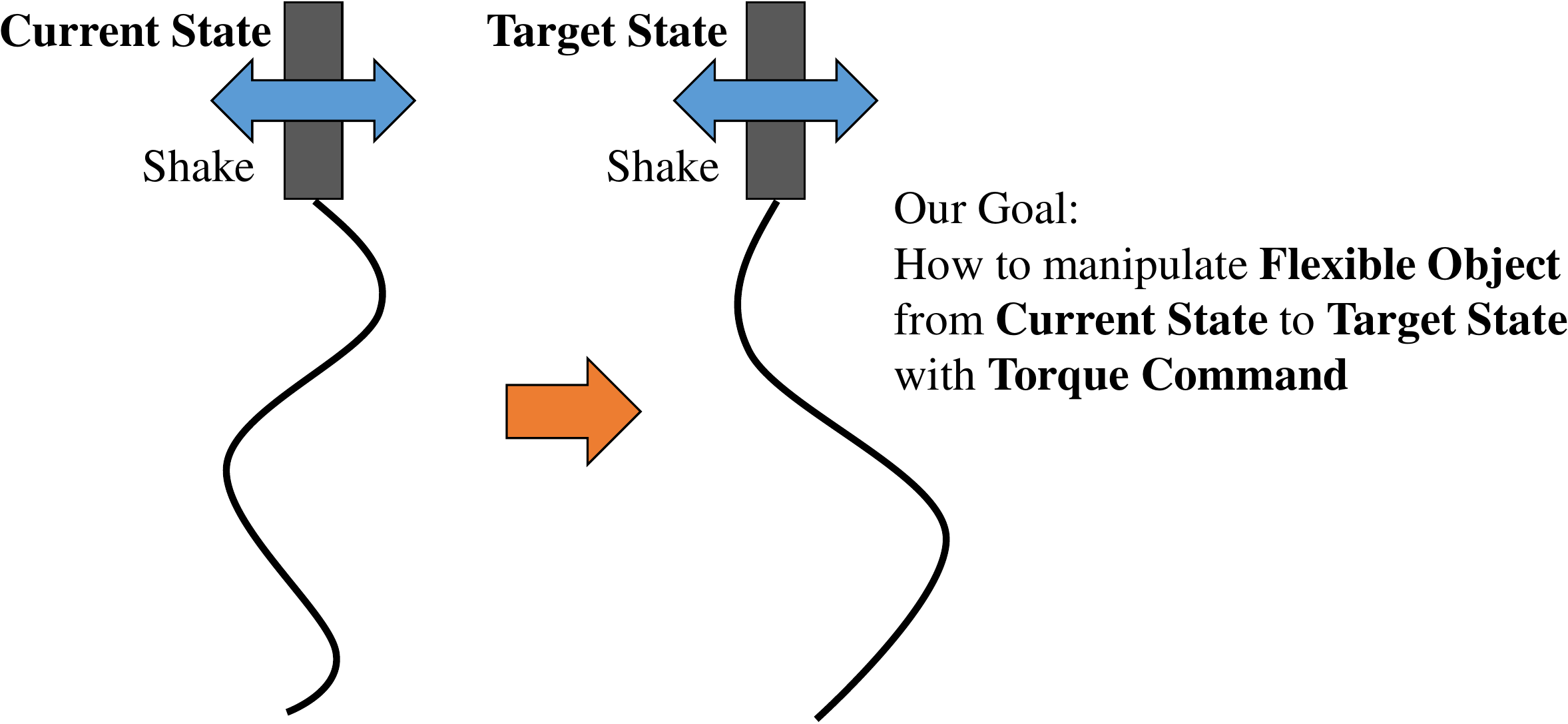}
  \caption{Our goal in this study.}
  \label{figure:motivation}
  \vspace{-1.3ex}
\end{figure}

\begin{figure}[b]
  \centering
  \includegraphics[width=1.0\columnwidth]{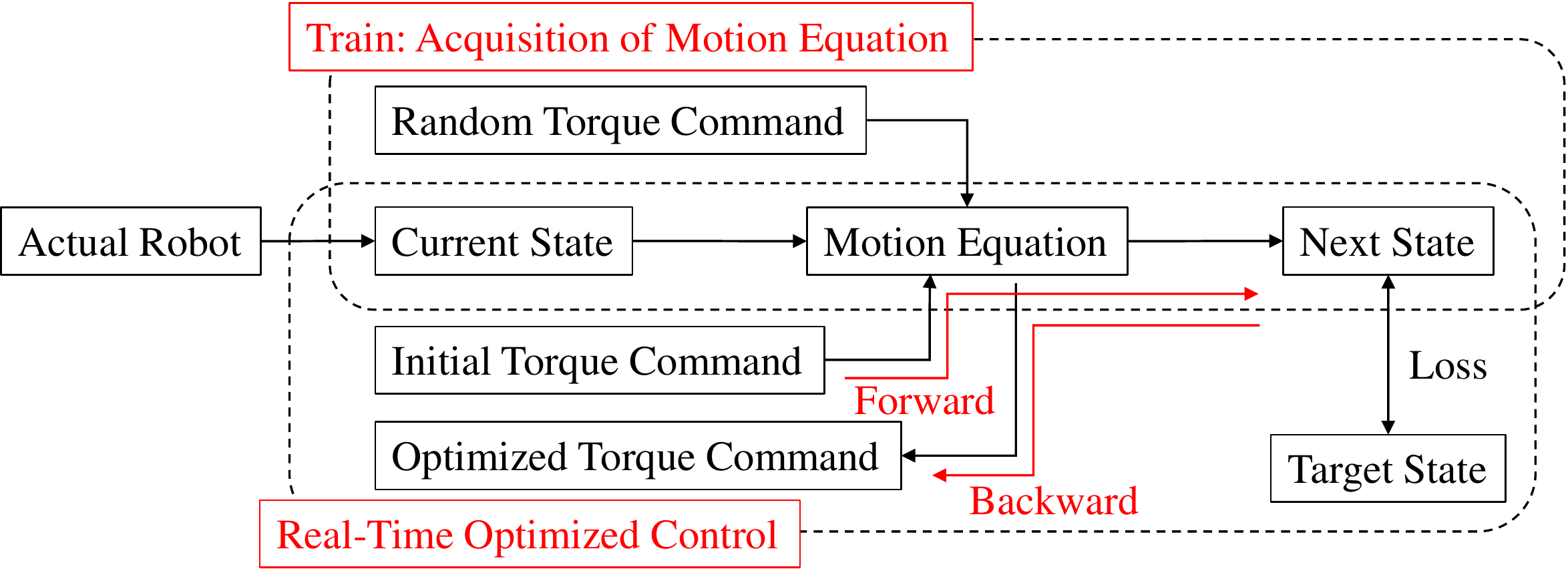}
  \caption{System overview of this study.}
  \label{figure:overview}
  \vspace{-1.3ex}
\end{figure}

\begin{figure*}[t]
  \centering
  \includegraphics[width=1.7\columnwidth]{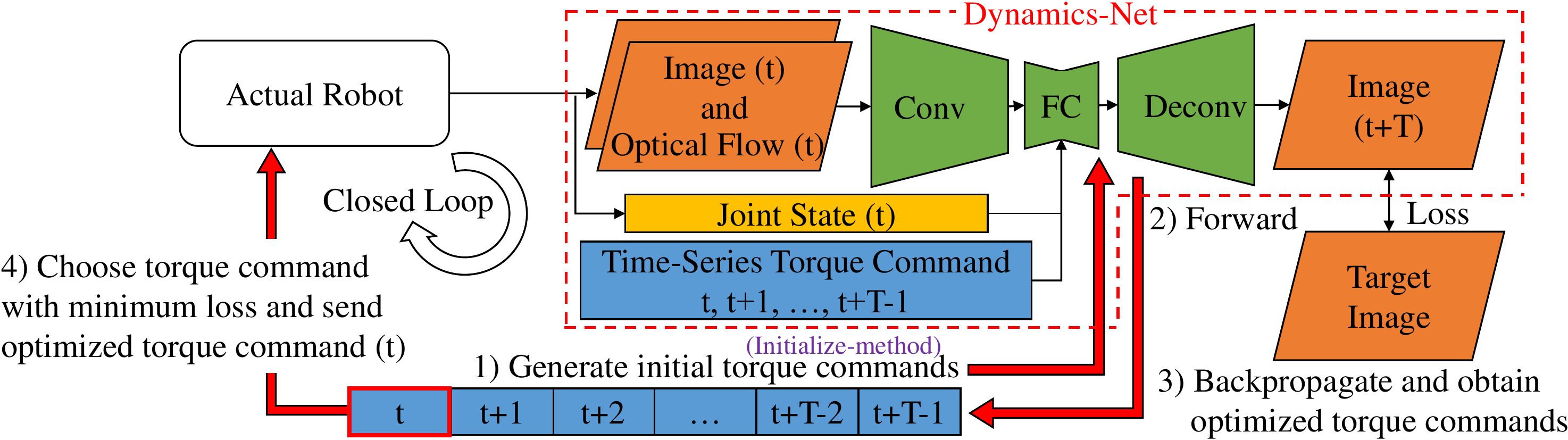}
  \caption{Network structure of Dynamics-Net and calculation of optimized time-series torque command.}
  \label{figure:dynamics-net}
  \vspace{-1.3ex}
\end{figure*}

  To realize real-time dynamic manipulation of flexible objects, we developed a system (\figref{figure:overview}) which has characteristics:
  \begin{itemize}
    \item Acquisition of flexible object motion equation model from motion data with random torque command using a deep neural network
    \item Calculation of optimized time-series torque command by backpropagation to the input torque command using the acquired motion equation model
    \item Generation of initial time-series torque commands before the backpropagation for real-time dynamic control
  \end{itemize}
  By using this system, we can move the flexible object from a current state to a target state dynamically with time-series torque command.
  The detailed contributions of this study are as shown below.
  \begin{itemize}
    \item A network structure representing motion equation using vision image and time-series torque command
    \item Real-time calculation process of optimized time-series torque command
    \item Comparison of task realization with changes of parameters in several experiments
  \end{itemize}

  In the following sections, at first, we introduce Dynamics-Net, which represents the dynamic motion equation of a target object.
  Then, we explain the calculation method of optimized time-series torque command using this Dynamics-Net.
  Finally, we verify the effectiveness of this method by manipulation experiments of a rigid object, a flexible object with and without environmental contact, and a cloth.

\section{Dynamics-Net and Our System} \label{sec:dynamics-net}
\subsection{Dynamics-Net} \label{subsec:network-structure}
  A general motion equation of robot manipulator can be formulated as,
  \begin{align}
    M(\bm{\theta})\ddot{\bm{\theta}} + \bm{c}(\bm{\theta}, \dot{\bm{\theta}}) + \bm{g}(\bm{\theta}) = \bm{\tau} \label{eq:motion-equation}
  \end{align}
  where $\bm{\theta}, \dot{\bm{\theta}}, \ddot{\bm{\theta}}$ are joint position, velocity, and acceleration respectively, $\bm{\tau}$ is joint torque, $M$ is an inertial matrix, $\bm{c}$ expresses centrifugal force, Coriolis force, viscous friction, etc., and $\bm{g}$ is gravity torque.
  When regarding the structure of flexible objects as under-actuated multi-link structures, we can represent the motion equation of flexible objects as \equref{eq:motion-equation}.
  This equation means that when we apply torque command $\bm{\tau}$ to the object with its current posture $\bm{\theta}$ and velocity $\dot{\bm{\theta}}$, we can obtain acceleration $\ddot{\bm{\theta}}$ at the next frame.
  Then, we can know the posture one frame or $T$ frames ahead, by integrating the acceleration.
  We represent the postures as images, since they can easily handle flexible objects, which have infinite degrees of freedom (DOFs).
  For the same reason, we use optical flow to represent the velocity.

  Dynamics-Net is a network representing this equation using a deep neural network as shown in \figref{figure:dynamics-net}.
  First, the convolution layers extract the current image feature from the image and optical flow of the target object.
  Second, the fully connected layers merge the current image feature, joint states (current position, velocity, and torque of each actuator), and time-series torque command until after $T$ frames.
  The frame length of time-series torque command $T$ is a hyperparameter, and we discuss it in \secref{sec:discussion}.
  Third, the deconvolution layers predict the final state, i.e., the image after $T$ frames, from the merged feature vector.

\begin{figure}[htb]
  \centering
  \includegraphics[width=0.8\columnwidth]{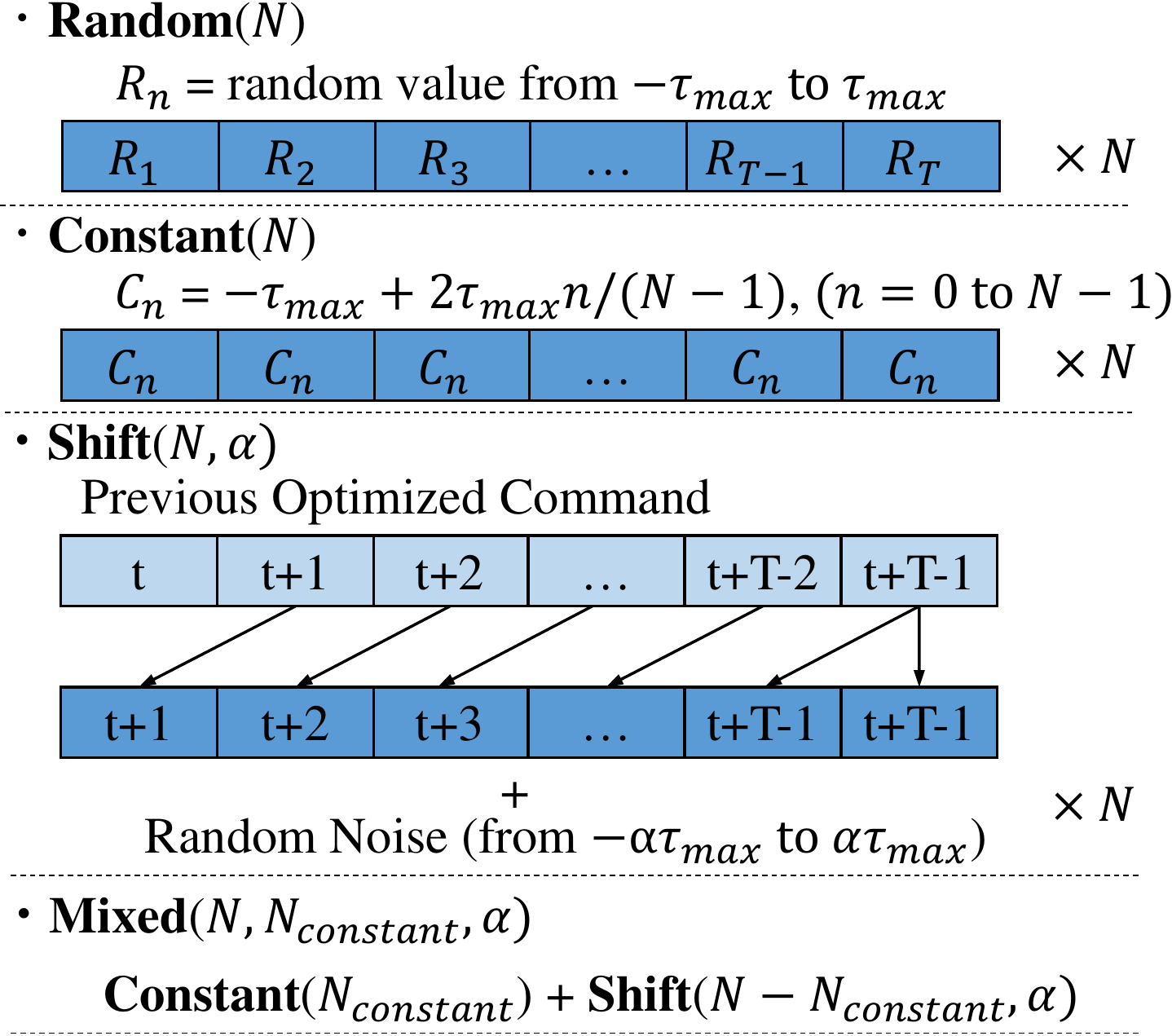}
  \caption{The method to generate initial time-series torque commands before the optimization.}
  \label{figure:input-method}
  \vspace{-1.3ex}
\end{figure}

\subsection{Calculation of Target Torque Command} \label{subsec:optimized-torque}
  We will explain how to calculate optimized time-series torque command to realize the target image of the flexible object using Dynamics-Net.
  The details are shown in \figref{figure:dynamics-net}, and the process is as follows.
  \begin{enumerate}
    \item Generate initial time-series torque commands before the optimization of the torque commands
    \item Obtain the current image and optical flow from the actual robot, feed them into Dynamics-Net with the initial time-series torque commands, and calculate the losses between the predicted and target images
    \item Optimize the initial time-series torque commands by backpropagation of the losses
    \item Choose the optimized torque command with the minimum loss, and send the torque command of the current frame as the target torque command
  \end{enumerate}

  Next, we will explain the details of 1)--4).

  First, 1) is the most controversial point in this study.
  We express the maximum joint torque as $\tau_{max}$ and the minimum as $-\tau_{max}$.
  \figref{figure:input-method} shows several generation methods of time-series torque commands for Dynamics-Net, named Initialize-methods.
  We define 4-type Initialize-methods:
  \begin{itemize}
    \item \textbf{Random}($N$): generate $N$ random time-series torque commands within $[-\tau_{max}, \tau_{max}]$
    \item \textbf{Constant}($N$): divide the torque limit $[-\tau_{max}, \tau_{max}]$ equally into $N$ parts and generate $N$ constant time-series torque commands
    \item \textbf{Shift}($N, \alpha$): shift the previous optimized time-series torque command, add random noise within $[-\alpha\tau_{max}, \alpha\tau_{max}]$, and generate $N$ samples
    \item \textbf{Mixed}($N, N_{constant}, \alpha$): generate $N$ samples by adding \textbf{Constant}($N_{constant}$) and \textbf{Shift}($N-N_{constant}, \alpha$)
  \end{itemize}
  We set $N=10, \alpha=0.25, N_{constant}=3$.
  Because the parameter $N$ depends on the control frequency (30 Hz in this study) and computational resources, we do not discuss it much in this study.
  These Initialize-methods contribute the efficient search of time-series torque commands to realize the target image.
  Also, because several time-series torque commands realizing the same target image exist, we verify how these Initialize-methods contribute to the dynamic movement.

  Regarding 2), the loss function is important.
  We use binary images without color information as input and output images.
  We use the loss with decreased sensitivity to posture deviations by blurring the target image, as shown below,
  \begin{align}
    \bm{S}'_{target} &= 1.0-\textrm{tanh}(\beta\cdot\textrm{DT}(\bm{S}_{target})) \nonumber\\
    \textrm{Loss} &= \textrm{MSE}(\bm{S}_{predicted}-\bm{S}'_{target}) \label{eq:loss}
  \end{align}
  where $\bm{S}_{target}$ is the target image, $\bm{S}_{predicted}$ is the predicted image, $\textrm{tanh}$ is a hyperbolic tangent, $\textrm{DT (DistanceTransform)}$ is the image of distances to the nearest object pixel at each pixel, $\beta$ is a scaling factor (0.5 in this study), and $\textrm{MSE}$ is mean squared error.

  Regarding 3) and 4), devising of the backward for real-time control and optimization of time-series torque command are important.
  At first, for real-time control, we remove the useless backward calculation of convolution layers.
  In addition, because the current image and optical flow are the same at the current frame, we feed them into the convolution layers just once.
  Next, we optimize the time-series torque command by backpropagation \cite{rumelhart1986learning} like in \cite{tanaka2018emd, byravan2018se3},
  \begin{align}
    \bm{g} &= d\textrm{Loss}/d\bm{\tau}^{ts} \\
    \bm{\tau}^{ts}_{optimized} &= \bm{\tau}^{ts}_{initial} - \gamma\bm{g}/|\bm{g}|
  \end{align}
  where, $\bm{\tau}^{ts}$ expresses time-series torque command, $\bm{\tau}^{ts}_{initial}$ is the initial $\bm{\tau}^{ts}$, $\bm{\tau}^{ts}_{optimized}$ is the optimized $\bm{\tau}^{ts}$, $\gamma$ is a constant ($0.25\bm{\tau}_{max}$ in this study).
  $\bm{\tau}^{ts}_{initial}$ is constructed by $N$ samples, we choose one sample of $\bm{\tau}^{ts}_{initial}$ with the minimum loss after forwarding, and obtain $\bm{\tau}^{ts}_{optimized}$ by optimizing it.
  Then, we conduct the forwarding of Dynamics-Net again by setting $\bm{\tau}^{ts}_{initial}$ as $\bm{\tau}^{ts}_{optimized}$.
  As the target joint torque at this frame, we use $\bm{\tau}^{ts}_{optimized}(t)$ if the calculated loss is smaller than the loss before optimization; otherwise, we use $\bm{\tau}^{ts}_{initial}(t)$.

\subsection{Implementation Details} \label{subsec:whole-system}
  Regarding image processing, because we basically treat 2D movements, we preprocess the image by seven steps: crop, resize, background subtraction, blurring, thresholding, closing, and opening.
  We extract the region of interest (ROI) by cropping, and resize it to the size of $64\times64$.
  Regarding the optical flow, we make a 2-channel image of the x- and y-axis of the optical flow.

  Regarding the network structure of Dynamics-Net, as described in \subsecref{subsec:network-structure}, it consists of three blocks: Conv, FC, and Deconv.
  For Conv block, we use five convolution layers.
  Each layer has 4, 8, 16, 32, and 64 channels, respectively.
  The number of channels in the input layer is 3: image and optical flow (x and y).
  In all layers, the kernel size is $3\times3$, stride is $2\times2$, and padding is 1.
  In FC block with fully connected layers, the number of units are $128+M(3+T)$, 128, 128, 128, and 256, where $M$ is the number of actuators and $3+T$ means the joint states and time-series torque command of each actuator.
  Deconv block has deconvolution layers which are the same structures as Conv block.
  Batch normalization \cite{ioffe2015batch} is applied to each layer except the output layer.
  The activation function of each layer except the output layer is ReLU \cite{nair2010rectified}, and we normalize the output layer by sigmoid.

  We implemented our network by Chainer \cite{tokui2015chainer}, and conducted our whole processes without GPU.

\begin{figure}[t]
  \centering
  \includegraphics[width=1.0\columnwidth]{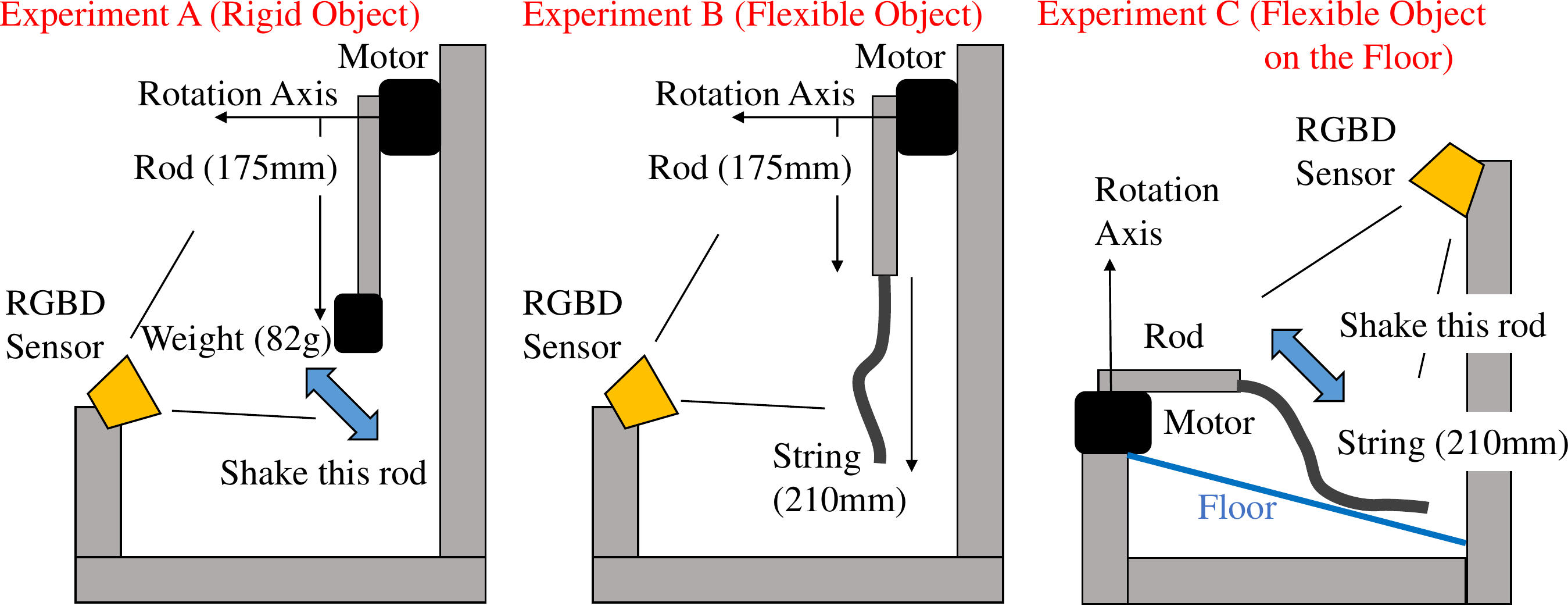}
  \caption{Details of experimental setup A (rigid object), B (flexible object), and C (flexible object on the floor).}
  \label{figure:system}
  \vspace{-1.3ex}
\end{figure}

\section{Experiments} \label{sec:experiments}
  We conducted four manipulation experiments (A--D).
  We evaluated our system quantitatively by manipulations of two fundamental objects: A) a rigid object and B) a flexible object.
  A) is the simplest setting bacause the target object has only one link, whereas the target object in B) has infinite DOFs and links.
  Also, we conducted two object manipulations in more practical situations: C) a flexible object on the floor and D) a cloth moving in 3D with 2 DOFs manipulator.
  In C), the object motion is affected by the environmental contact.
  In D), the motion equation is more complex and we have to control more than one actuators concurrently.

  \figref{figure:system} shows the experimental settings of A--C.
  We cover these experimental setups by a black curtain.
  The actuator is XM430-W350-R (Dynamixel motor), and RGBD sensor is D435 (Intel Realsense).
  In A), we added weight to the tip of the rod, so the rod cannot be raised slowly and needs to be moved dynamically.
  In C), the string moves on the floor, so there is a frictional force.
  We describe the setting of D) in \subsecref{subsec:handkerchief} separately.

  As an image similarity metric, we used chamfer distance $d_{chamfer}$ \cite{borgefors1988hierarchical} as shown below,
  \begin{align}
    d_{chamfer}(\bm{S}_1, \bm{S}_2)=\sum(\bm{S}_1\cdot\textrm{DT}(\bm{S}_2)+\bm{S}_2\cdot\textrm{DT}(\bm{S}_1))\label{eq:eval}
  \end{align}
  where the unit of $d_{chamfer}$ is px.
  We set the current image as $\bm{S}_1$ and the target image as $\bm{S}_2$.
  The target image can change over time, but we set a constant image as the target image in this study.

\begin{figure}[htb]
  \centering
  \includegraphics[width=1.0\columnwidth]{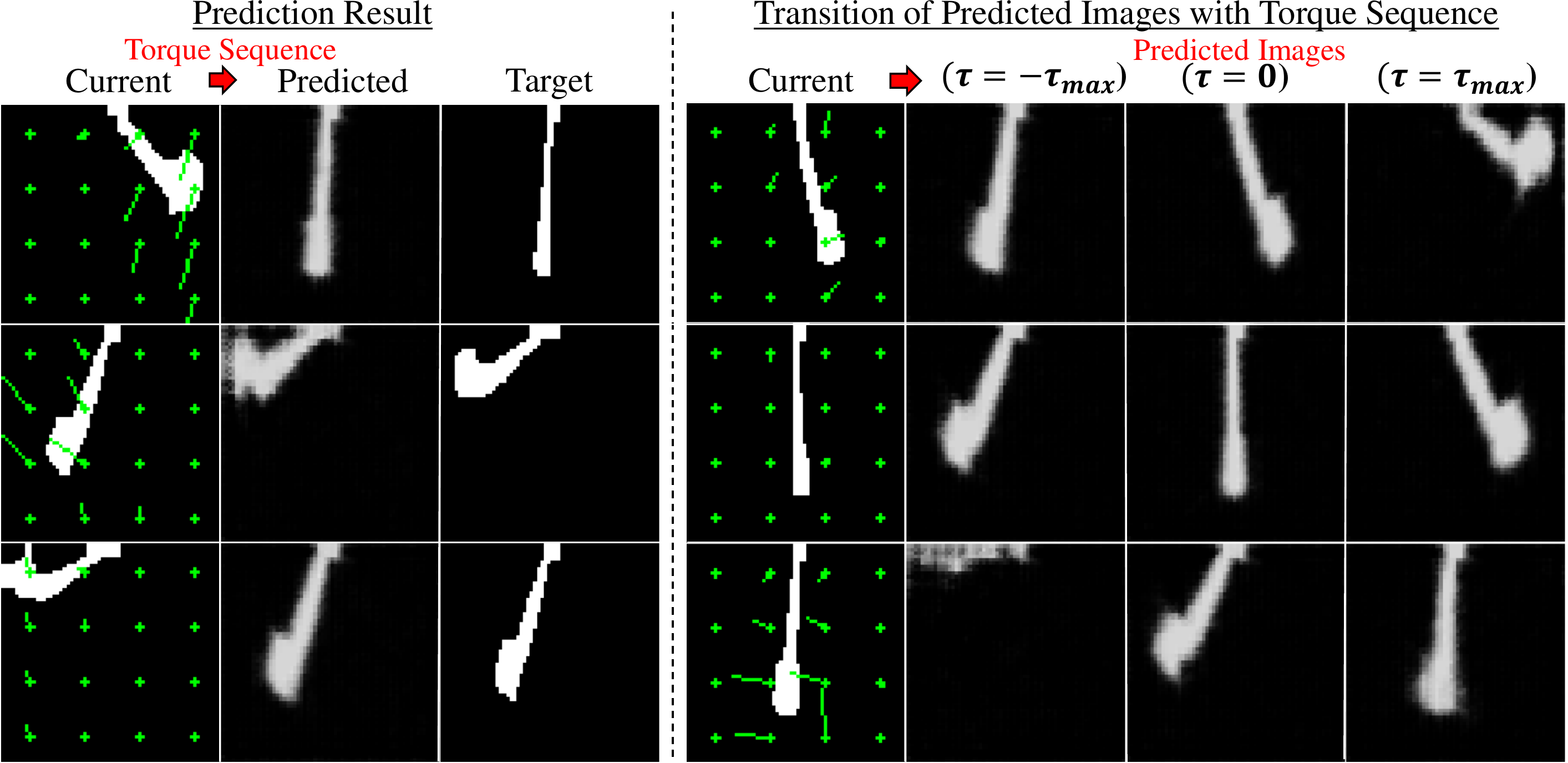}
  \caption{Prediction result using the trained Dynamics-Net and transition of predicted images with time-series torque command, in rigid object experiment.}
  \label{figure:rigid-summarize}
  \vspace{-1.3ex}
\end{figure}

\subsection{Rigid Object Manipulation} \label{subsec:rigid-body}
  At first, we collected $10,800$ frames of motion data of rigid object manipulation by random torque commands for 6 minutes, where we set $\tau_{max}=0.2[\textrm{Nm}]$ and $d\tau_{max} = 0.1[\textrm{Nm/frame}]$.
  This data collection does not need any annotation and we only need to wait for a short time.
  Then, we trained Dynamics-Net using the data.
  We show the prediction results in the left figure of \figref{figure:rigid-summarize}, where $T = 10$ and the green lines express optical flow thinning out for visualization.
  We can see that Dynamics-Net can predict images after 10 frames properly and the predicted images are blurred due to the loss function of \equref{eq:loss} which blurs target images.
  Next, we show the transition of predicted images by changing the time-series torque command $\bm{\tau}^{ts} = -\bm{\tau}^{ts}_{max}, \bm{0}, \bm{\tau}^{ts}_{max}$, respectively, in the right figure of \figref{figure:rigid-summarize}.
  We can see that the predicted images transition by the change of the torque command and they are reasonable from the current image and optical flow.

\begin{figure*}[t]
  \centering
  \includegraphics[width=1.8\columnwidth]{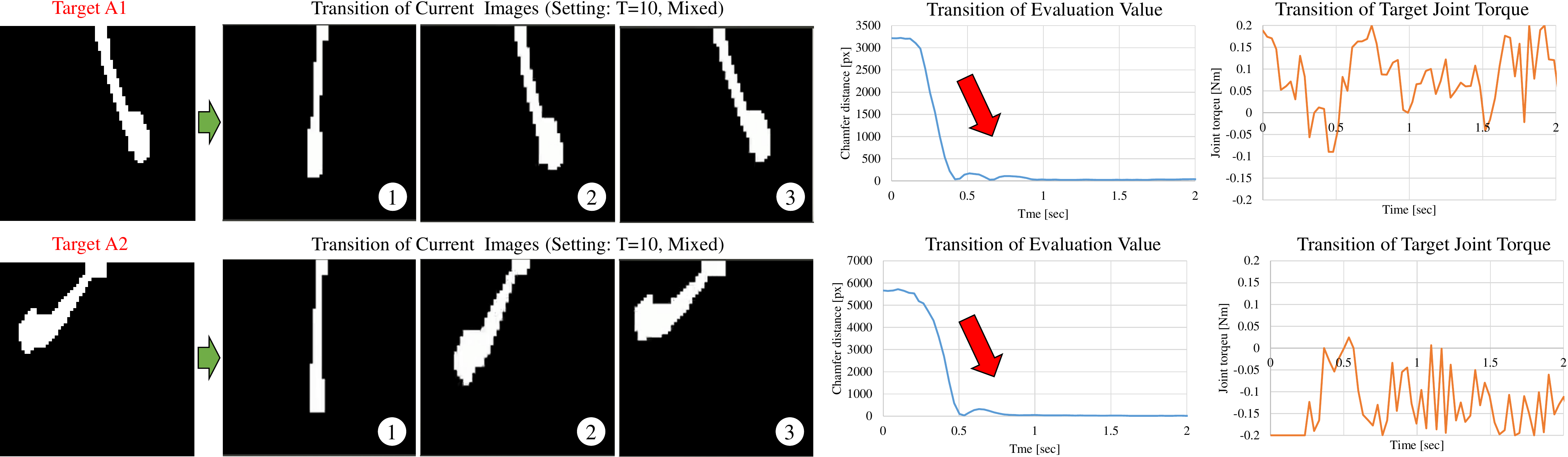}
  \caption{Rigid object manipulation: the transition of current images, evaluation value, and target joint torque.}
  \label{figure:rigid-actual-experiment}
  \vspace{-1.3ex}
\end{figure*}

\begin{figure*}[t]
  \centering
  \includegraphics[width=2.0\columnwidth]{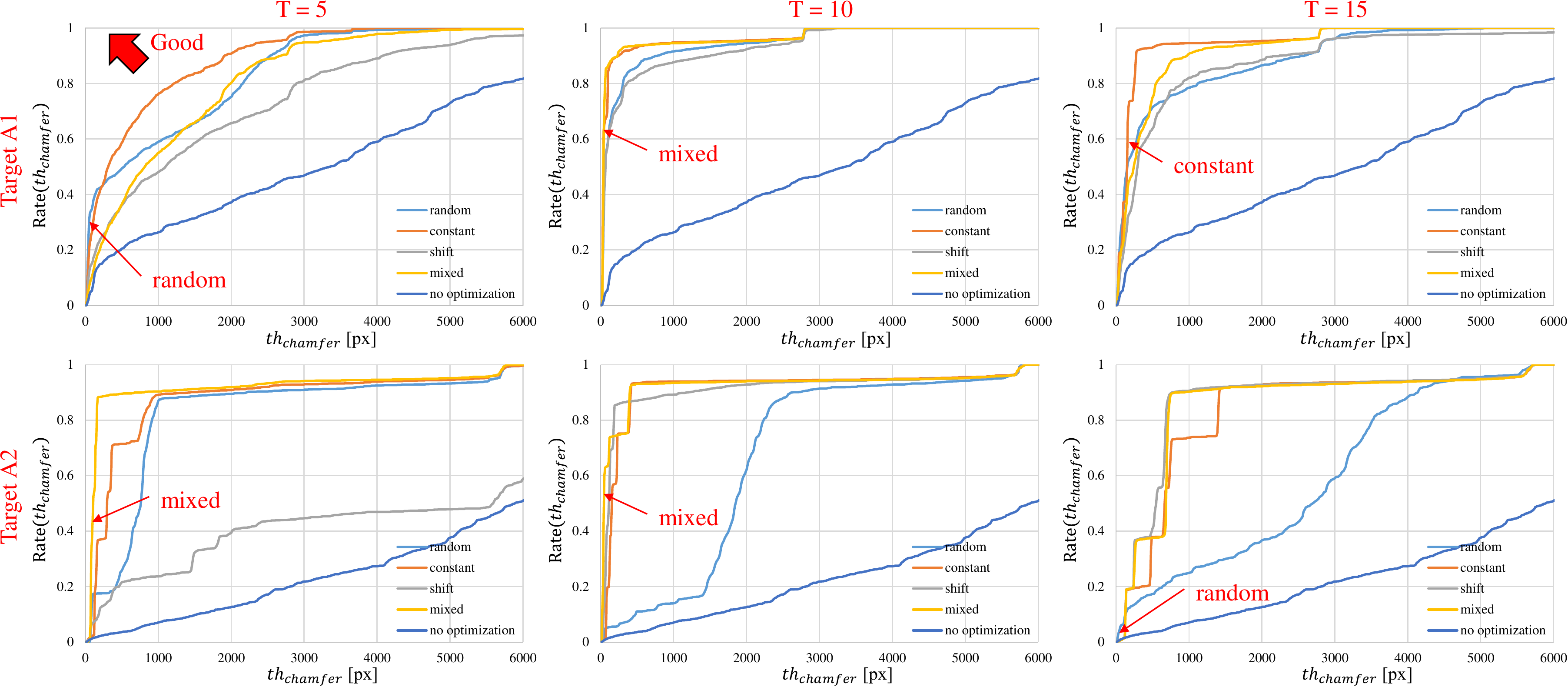}
  \caption{The relationship between $th_{chamfer}$ and rate of frames below the $th_{chamfer}$: $\textrm{Rate}(th_{chamfer})$, when changing Initialize-methods, $T$, and target images, in rigid object manipulation.}
  \label{figure:rigid-input-evaluation}
  \vspace{-1.3ex}
\end{figure*}

  Finally, we conducted control experiments of a rigid object to realize the target image using Dynamics-Net.
  First, we show how the target images were realized by the proposed method in \figref{figure:rigid-actual-experiment}.
  We prepared Target A1 and Target A2 as target images.
  We set $T=10$ and used \textbf{Mixed} as Initialize-method.
  We can see the evaluation value $d_{chamfer}$ converged rapidly to about zero.
  Also, the target joint torque vibrated to keep the target image after the convergence.
  Second, we show the comparison of target image realization when changing $T$ to $5, 10, 15$ and Initialize-method to \textbf{Random}, \textbf{Constant}, \textbf{Shift}, and \textbf{Mixed}, in \figref{figure:rigid-input-evaluation}.
  We conducted a 10-second experiment 5 times regarding each parameter from the state with the rod taken down, and measured the target image realization.
  The horizontal axis of each graph expresses the threshold of $d_{chamfer}$ ($th_{chamfer}$), and the vertical axis expresses the rate of frames below the $th_{chamfer}$ (we express it $\textrm{Rate}(th_{chamfer})$).
  Therefore, a good parameter has higher $\textrm{Rate}(th_{chamfer})$ when $th_{chamfer}$ is low.
  \textbf{No Optimization} refers to the line of the graph using random torque control without any optimization.
  In \figref{figure:rigid-input-evaluation}, with all parameters, our framework realized target images better than \textbf{No Optimization}, and $T=10$ is the best.
  Also, when $T=10$, \textbf{Mixed} can realize target images the best, and the second best is \textbf{Constant}.

\subsection{Flexible Object Manipulation} \label{subsec:flexible-air}
\begin{figure}[htb]
  \centering
  \includegraphics[width=1.0\columnwidth]{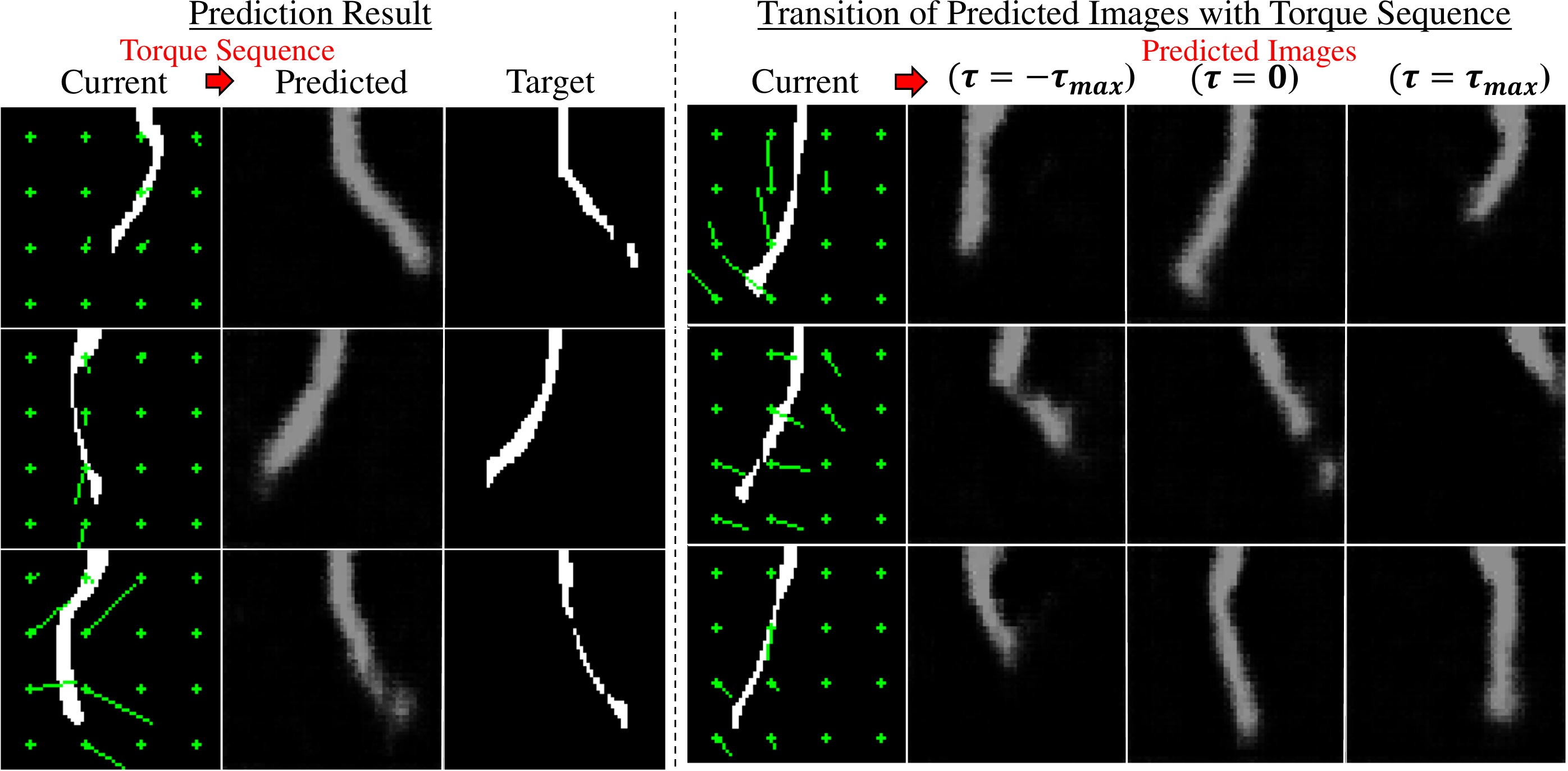}
  \caption{Prediction result using the trained Dynamics-Net and transition of predicted images with time-series torque command, in flexible object experiment.}
  \label{figure:string-air-summarize}
  \vspace{-1.3ex}
\end{figure}

\begin{figure*}[t]
  \centering
  \includegraphics[width=1.8\columnwidth]{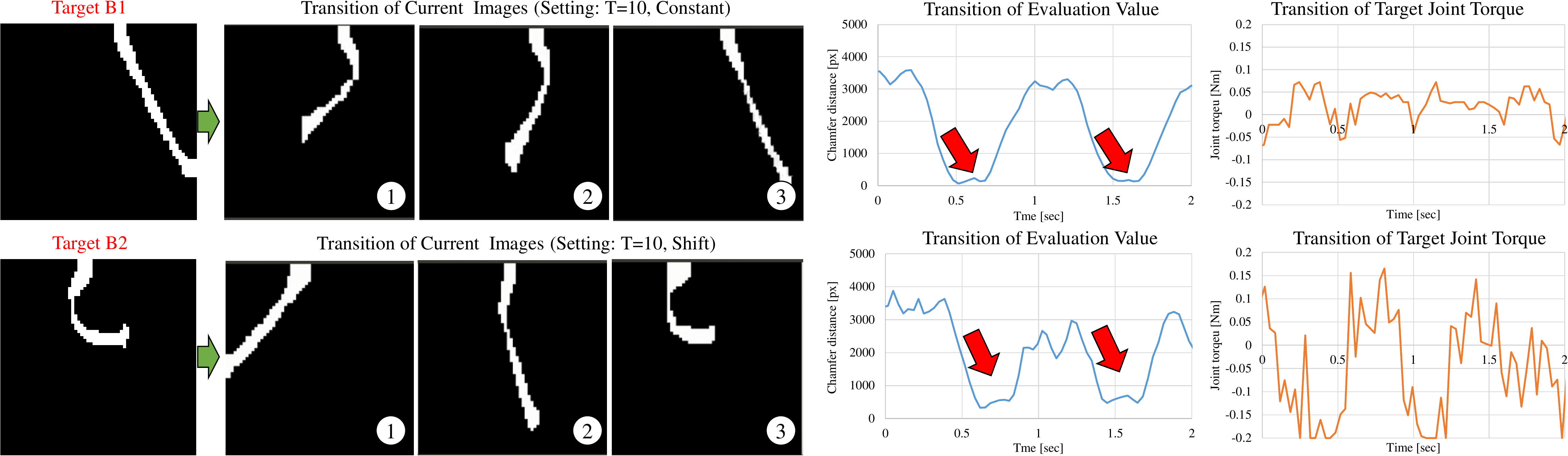}
  \caption{Flexible object manipulation: the transition of current image, evaluation value, and target joint torque.}
  \label{figure:string-air-actual-experiment}
  \vspace{-1.3ex}
\end{figure*}

\begin{figure*}[t]
  \centering
  \includegraphics[width=2.0\columnwidth]{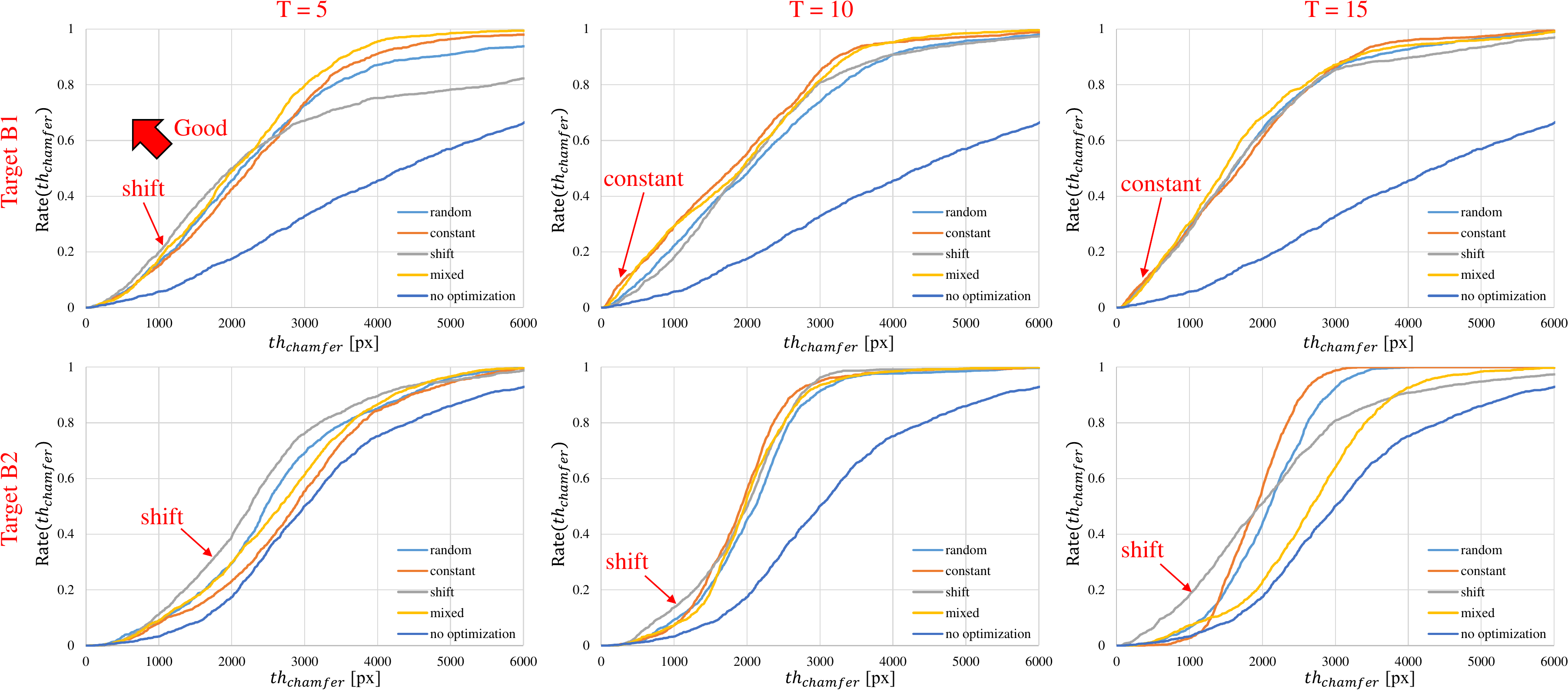}
  \caption{The relationship between $th_{chamfer}$ and rate of frames below the $th_{chamfer}$: $\textrm{Rate}(th_{chamfer})$, when changing Initialize-methods, $T$, and target images, in flexible object manipulation.}
  \label{figure:string-air-input-evaluation}
  \vspace{-1.3ex}
\end{figure*}

  We conducted a flexible object manipulation experiment, but the basic setting is the same as the rigid object manipulation.
  As with the rigid object experiment, we trained Dynamics-Net from $10,800$ motion data of random torque control of a flexible object.
  We show the prediction results in the left figure of \figref{figure:string-air-summarize}, and the transition of predicted images by changing torque command in the right figure of \figref{figure:string-air-summarize}.
  We can see that Dynamics-Net can predict images after 10 frames properly and the transition of predicted images is also reasonable.
  Then, we prepared Target B1 and B2 as target images, and show how target images were realized in \figref{figure:string-air-actual-experiment}.
  We set $T=10$ and used \textbf{Constant} regarding Target B1 and \textbf{Shift} regarding Target B2.
  We can see the evaluation value vibrated slowly, because we set images which cannot stand still as target images.
  Also, we show the comparison of target image realization when changing $T$ to $5, 10, 15$ and Initialize-method to \textbf{Random}, \textbf{Constant}, \textbf{Shift}, and \textbf{Mixed}, in \figref{figure:string-air-input-evaluation}.
  With all parameters, our framework realized target images better than \textbf{No Optimization}, and $T=10,15$ are better than $T=5$.
  Also, \textbf{Constant} and \textbf{Mixed} can realize Target B1 the best, and \textbf{Shift} can realize Target B2 the best.

\subsection{Flexible Object Manipulation on the Floor} \label{subsec:flexible-floor}
\begin{figure}[t]
  \centering
  \includegraphics[width=1.0\columnwidth]{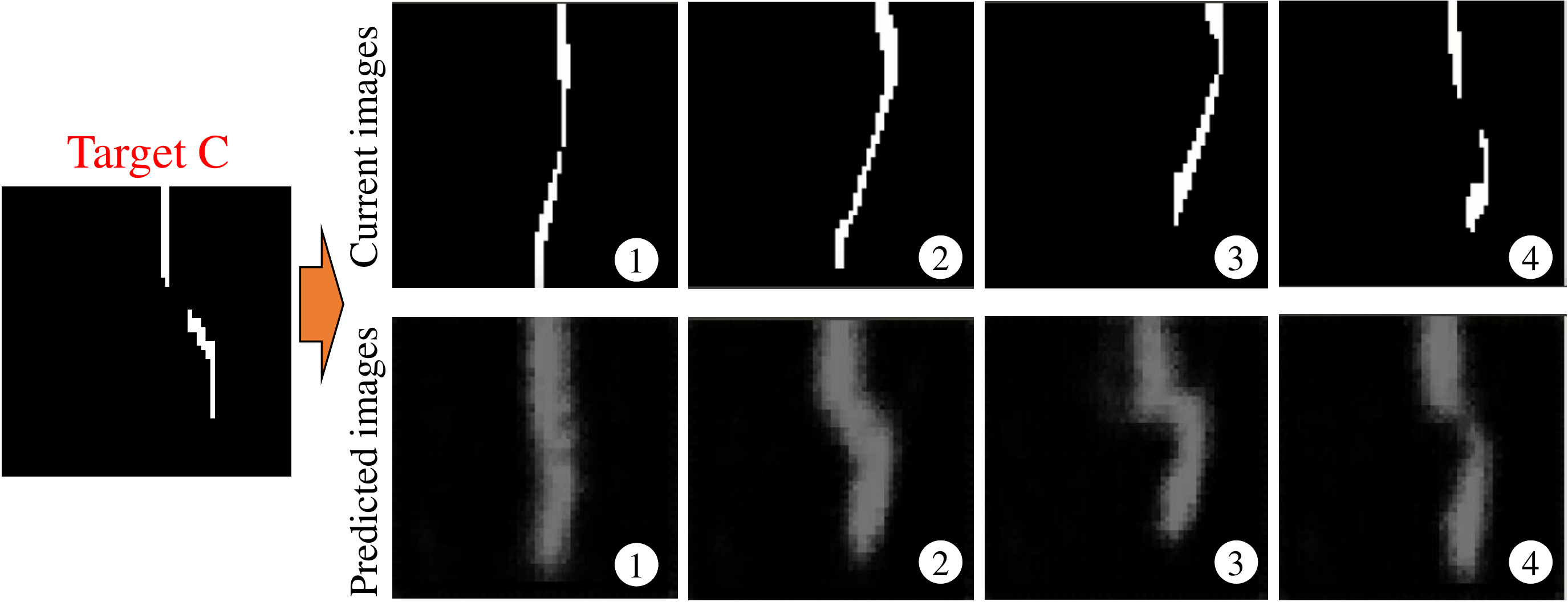}
  \caption{Flexible object manipulation on the floor: the transition of current and predicted images.}
  \label{figure:string-floor-actual-experiment}
  \vspace{-1.3ex}
\end{figure}

  In this experiment, because environmental contact is added, motion generation using physics-based modeling is more difficult due to the difficulty of friction modeling.
  We prepared Target C as a target image, and set $T=10$ and \textbf{Mixed} as Initialize-method.
  The result is shown in \figref{figure:string-floor-actual-experiment}.
  The predicted images realized approximately the Target C, and the current images realized approximately the Target C afterwards.
  Therefore, Dynamics-Net can even acquire motion equation model with environmental forces such as frictional force.

\subsection{Cloth Manipulation} \label{subsec:handkerchief}
\begin{figure}[t!]
  \centering
  \includegraphics[width=1.0\columnwidth]{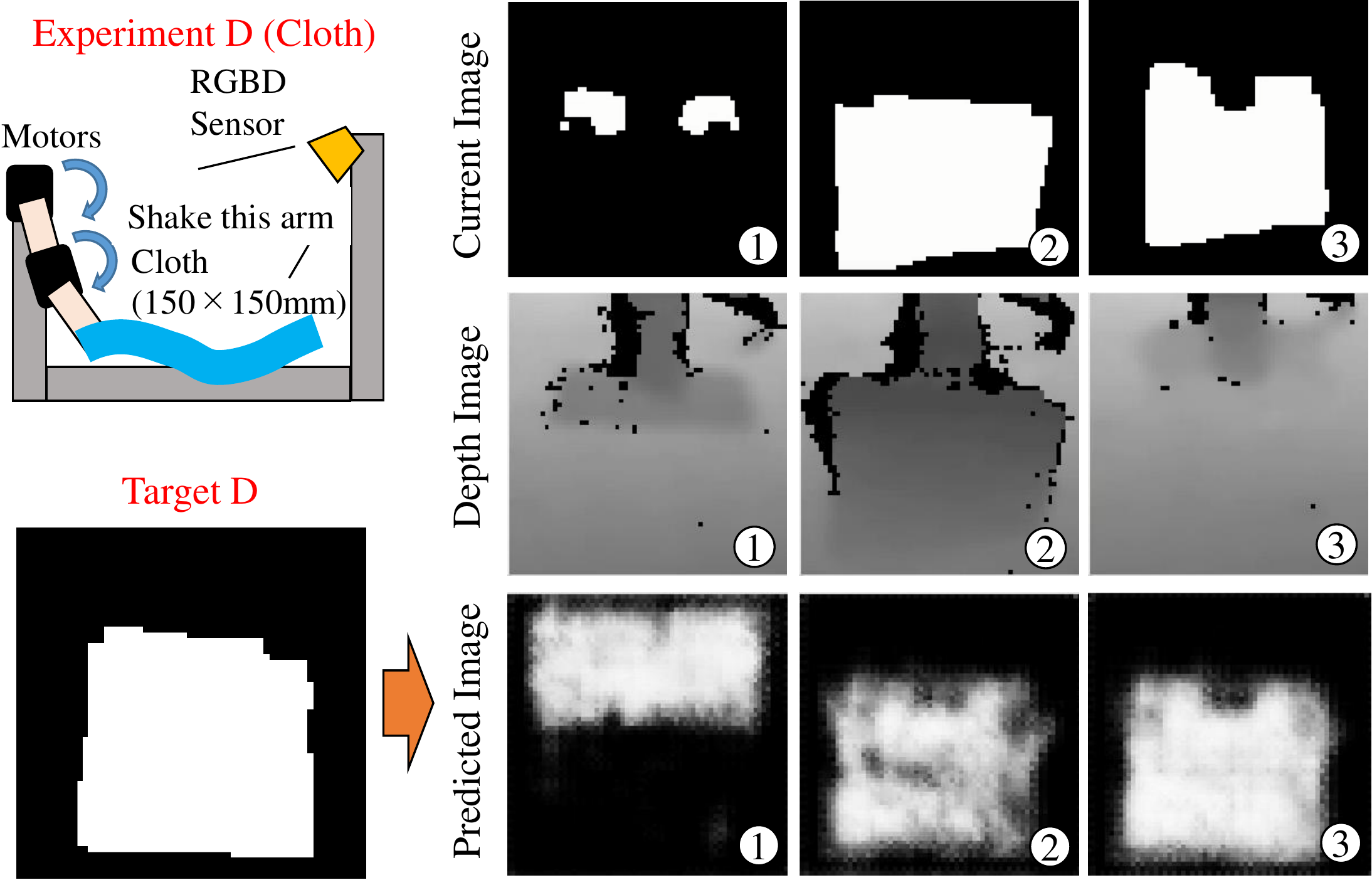}
  \caption{Cloth manipulation in 3D: the experimental setup, and the transition of current, depth, predicted images.}
  \label{figure:handkerchief-actual-experiment}
  \vspace{-1.3ex}
\end{figure}

  We conducted cloth manipulation in 3D with a robot having 2 DOFs.
  The experimental setup is shown in the upper left figure of \figref{figure:handkerchief-actual-experiment}.
  Although we have handled 2D movements so far, to consider 3D movements, we changed the input layer of Dynamics-Net to 4 channels: current image, optical flow (x and y), and depth image.
  The values of the depth image are cropped and normalized to a value from 0 to 1.
  Ideally, we must add the optical flow in the z-axis to the input layer and the depth image to the output layer, but we did not add these changes due to the difficulty of the loss function setting and increased computational complexity.
  The actuators are XM430-W210-R (Dynamixel motor).
  We set $\tau_{max}=0.3[\textrm{Nm}]$, $T=10$, and \textbf{Mixed} as Initialize-method.
  We prepared Target D as a target image, and the folded cloth was unfolded properly by the proposed system, as shown in the right figure of \figref{figure:handkerchief-actual-experiment}.
  We show the feasibility to realize flexible object manipulation in 3D.

\section{Discussion} \label{sec:discussion}
  First, regarding the rigid object experiment, the results show that the best is \textbf{Mixed} and the second best is \textbf{Constant}, but regarding the flexible object experiment, \textbf{Shift} is the best, especially in Target B2.
  From these results, we can consider that \textbf{Constant} is good at realizing static movements because it provides constant torques, and \textbf{Shift} is good at realizing dynamic movements because it considers the time-series torque transition.
  In the rigid object experiment, we assume that \textbf{Shift} is effective approaching the target image, and \textbf{Constant} is effective at approximately the target image, so the target image was realized the best by using \textbf{Mixed} mixing \textbf{Constant} with \textbf{Shift}.
  On the other hand, in the flexible object experiment, especially Target B2 cannot be realized by static movements, so \textbf{Mixed} is not good for this target image, and \textbf{Shift}, which considers only time-series torque command transition, is the best.

  Second, from the experimental results, we can see that $T$ should not be too small or too large.
  This is because we cannot realize target images far away when $T$ is too small, and we cannot obtain optimized torque command correctly, due to the difficulty of predicting the distant future, when $T$ is too large.

  Third, the Dynamics-Net we implemented in this study is one of the simplest structures representing the motion equation, and we can consider other various structures.
  For example, we can insert recurrent structures like LSTM \cite{hochreiter1997long}, or make the network structure lighter by using features extracted from AutoEncoder \cite{hinton2006reducing}.
  Also, we can adopt $T$ as a network input to use it as a variable, or output the images not only at $t+T$ but also at $[t+1, t+T]$.
  These changes may be able to consider the transition of current images and remove the hyperparameter $T$.

\section{CONCLUSION} \label{sec:conclusion}
  In this study, we proposed an acquisition method of a motion equation model of flexible object manipulation with torque command by Dynamics-Net and a calculation method of optimized time-series torque command to realize a target state.
  Dynamics-Net works as the motion equation of flexible object manipulation whose inputs are the current image, optical flow, joint states, and time-series torque command, and whose output is the predicted image after $T$ frames.
  We also investigated the calculation method of optimized time-series torque command by backpropagation to inputs and the generation method of initial time-series torque commands for dynamic real-time control.
  Then we validated the effectiveness by experiments using a rigid object, a flexible object, a flexible object on the floor, and a cloth.
  In future works, we would like to validate the effectiveness of Dynamics-Net when using a multi-DOFs manipulator, and apply this method to more practical tasks.

\bibliographystyle{junsrt}
\bibliography{main}

\end{document}